\documentclass[10pt,twocolumn,letterpaper]{article}
\pdfoutput=1

\usepackage{cvpr}
\usepackage{times}
\usepackage{epsfig}
\usepackage{graphicx}
\usepackage{amsmath}
\usepackage{amssymb}
\usepackage{float}
\usepackage{setspace}
\usepackage{enumitem}

\usepackage{amsthm}
\usepackage{authblk}

\usepackage{hyperref}
\hypersetup{breaklinks=true,colorlinks}
\usepackage[
  separate-uncertainty = true,
  multi-part-units = repeat
]{siunitx}

\cvprfinalcopy % *** Uncomment this line for the final submission

\def\cvprPaperID{11} % *** Enter the CVPR Paper ID here

\ifcvprfinal\fi
\begin{document}

\setlength{\jot}{3pt}

%%%%%%%%% TITLE
\title{SegAttnGAN: Text to Image Generation with Segmentation Attention}

\author[1]{Yuchuan Gou}
\author[2]{Qiancheng Wu}
\author[1]{Minghao Li}
\author[1]{Bo Gong}
\author[1]{Mei Han}

\affil[1]{PAII Inc. \{\tt gouyuchuan355, liminghao058, gongbo173, hanmei613\}@paii-labs.com}
\affil[2]{University of California, Berkeley \{\tt qcwu\}@berkeley.edu}

\maketitle
%\thispagestyle{empty}

%%%%%%%%% ABSTRACT
\begin{abstract}
In this paper, we propose a novel generative network (SegAttnGAN) that utilizes additional segmentation information for the text-to-image synthesis task.  As the segmentation data introduced to the model provides useful guidance on the generator training,  the proposed model can generate images with better realism quality and higher quantitative measures compared with the previous state-of-art methods. We achieved Inception Score of 4.84 on the CUB dataset ~\cite{WahCUB_200_2011} and 3.52 on the Oxford-102 dataset ~\cite{Nilsback08}. Besides, we tested the self-attention SegAttnGAN which uses generated segmentation data instead of masks from datasets for attention and achieved similar high-quality results, suggesting that our model can be adapted for the text-to-image synthesis task.
% In this paper, we propose a generative network (SegAttnGAN) with a novel segmentation attention model that utilizes segmentation information to constrain text-to-image GAN model training. The proposed model is able to generate images with better realism quality and higher quantitative measures. We achieved the state-of-art Inception Score of 4.84 on the CUB dataset, boosting the baseline model (AttnGAN) by 11 percent. In addition, we test the self-attention SegAttnGAN using generated segmentation data for attention and achieved similar high-quality results, suggesting that our model can be adapted for the text-to-image task.
\end{abstract}

%%%%%%%%% BODY TEXT
\section{Introduction}
The task of generating high fidelity, realistic-looking images based on semantic description is central to many applications. A lot of research has been focused on the text-to-image synthesis task, which takes in natural language descriptions to generate images matching the text. 

\begin{figure}[H]
\begin{center}
    \includegraphics[width=\linewidth]{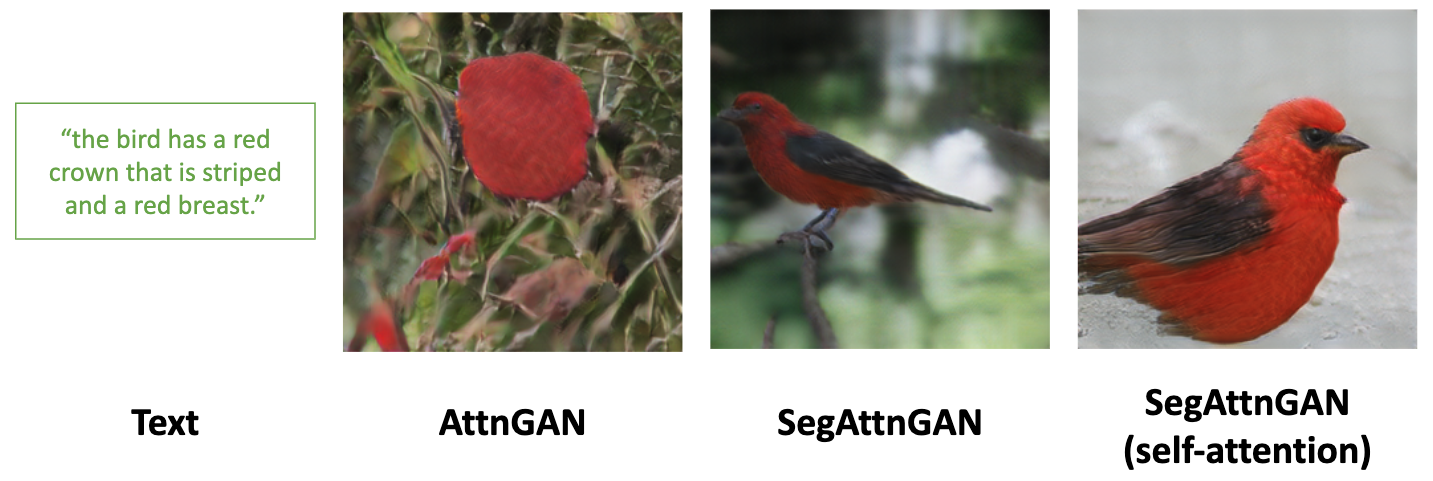}
\end{center}
  \caption{Sample results generated from AttnGAN~\cite{Tao18attngan}, our proposed SegAttnGAN and self-attention SegAttnGAN.}
\label{fig:intro_result}
\end{figure}

Many models for this task use generative adversarial networks (GANs) ~\cite{han2017stackgan, Tao18attngan, Han17stackgan2, reed2016generative, qiao2019mirrorgan}, conditioned on the text input rather than Gaussian noise for image generation. Although models like ~\cite{Tao18attngan} achieve satisfactory visual quality while maintaining the image-text consistency, there is little control over the layout of the generated images except for specific keywords which uniquely constrain the shape of the objects. Frequently these models would generate objects with deformed shapes or images with unrealistic layouts (see Figures \ref{fig:intro_result} and \ref{fig:result1}).

Recent work in ~\cite{park2019SPADE} has shown that decent results can be achieved for image synthesis task when spatial attention from segmentation data is utilized to guide image generation. To solve the deformed shapes and unrealistic layouts problems, we designed SegAttnGAN, which utilizes the segmentation to add global spatial attention in addition to text input. We hope that the spatial information would regulate the layout of generated images thus create more realistic images. Experimentation has shown promising results when additional segmentation information is used to guide image generation.
% Specifically, we propose a model that takes text description as well as segmentation masks as input. The model employs a coarse-to-fine architecture and the segmentation map is added as attention after each upsampling layer. 
Our contributions can be summarized as the following:
\begin{enumerate}[parsep=10pt,itemsep=-10pt]
\item 
We proposed a novel generative network that uses both text and spatial attention to generate realistic images. 
%Based on the basic network structure from AttnGAN, we designed SegAttnGAN with segmentation input in a spatially adaptive way in each upsampling layer in the generator.
\item
We verified that the addition of spatial attention mechanism to GAN could substantially increase visual realism by regulating object shapes and image layouts.
\item
We built a self-attention network to generate segmentation masks first and then use it for image generation. Based on the qualitative results, self-attention model can also constrain the object shapes very well.
%We showed that model-generated segmentation maps from text description could be used as the additional input for the spatial attention module without compromising output quality. This finding allows us to build a 2-stage network that generates predicted segmentation maps from text input and then feeds into SegAttnGAN model for image generation, making our model applicable to the text-to-image task. 
\end{enumerate}

%------------------------------------------------------------------------
\section{Related Work}
As text-to-image synthesis played an important role in many applications, different techniques have been proposed for text-to-image synthesis task. Reed \etal ~\cite{reedpixelcnn} utilized PixelCNN to generate image from text description. Mansimov \etal ~\cite{mansimov16_text2image} proposed a model iteratively draws patches on a canvas, while attending to the relevant words in the description and Nguyen \etal ~\cite{nguyen} used an approximate Langevin sampling approach to generate image conditioned on text.

Since Goodfellow \etal ~\cite{goodfellow} introduced Generative Adversarial Networks (GANs), extensive research has been conducted on image generation task with different types of GANs and high-quality results have been achieved ~\cite{radford2015, conditionalgan, pggan, biggan, stylegan, zhu2017unpaired, pix2pix2017, park2019SPADE}. At the same time, researchers have also started to apply GAN techniques on text-to-image synthesis tasks. Reed \etal ~\cite{reed2016generative} proposed a conditional GAN to generate images of birds and flowers from detailed text descriptions and in ~\cite{reed2016learning} they added object location control to the conditional GAN. Zhang \etal ~\cite{han2017stackgan} proposed StackGAN to generate images from text. StackGAN consists of Stage-I and Stage-II GANs where the Stage-I GAN generates low-resolution images and the Stage-II GAN generates high-resolution images. Compared with StackGAN which is conditioned on sentence level, AttnGAN proposed by Xu \etal ~\cite{Tao18attngan} develops conditioning on both sentence level and word level aiming at generating fine-grained high-quality images from text descriptions. Zhang \etal ~\cite{zhang2018hdgan} proposed a hierarchically-nested GAN for text-to-image synthesis. Qiao \etal \cite{qiao2019mirrorgan} proposed MirrorGAN in order to achieve both visual realism and semantic consistency. Hong \etal ~\cite{hong18} and Li \etal ~\cite{objgan19} are both concentrating on text-to-image synthesis task in a coarse-to-fine way. But their focus is the word embedding module and object-level discrimination by designing a bidirectional LSTM on either global or object level. While our focus lies on the generator with attention mechanism to effectively constrain the object boundary given segmentation maps.

Semantic information provides useful guidance in image generation. It has been introduced as input in different formats. Works in ~\cite{huang2018munit, pix2pix2017, zhu2017multimodal} used edge map as guidance in image to image translation. Karacan \etal ~\cite{Karacan2016LearningTG} and Park \etal ~\cite{park2019SPADE} used semantic layout as guidance in image generation. In ~\cite{gu2019maskguided}, facial masks have been provided as guidance to generate faces. Our work differs from these works as we apply semantic masks in text-to-image synthesis task while theirs are dealing with image-to-image translation or image generation.

\section{SegAttnGAN for text-to-image synthesis}
% As shown in Figure \ref{fig:generator}, the proposed model takes text description and segmentation mask as inputs. And we also proposed a self-attention SegAttnGAN which generating segmentation for its attention.

\begin{figure*}[t]
\begin{center}
    \includegraphics[width=0.9\linewidth]{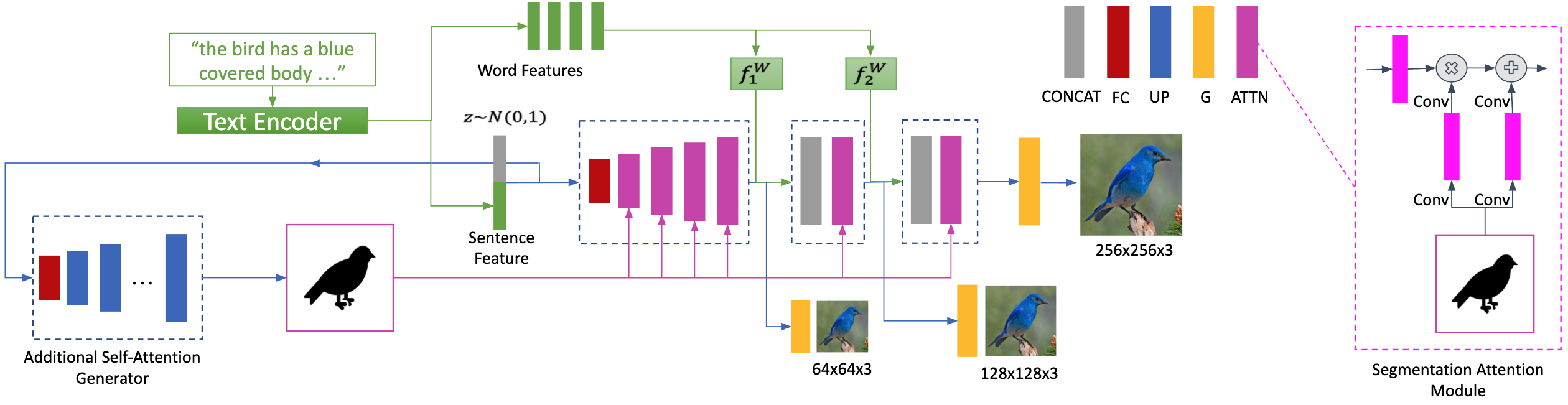}\end{center}
  \caption{Our proposed SegAttnGAN architecture. When the self-sttention generator is included, it becomes self-attention SegAttnGAN.}
\label{fig:generator}
\end{figure*}
\subsection{SegAttnGAN architecture}
Text-to-image generation models usually encode the whole sentence text description into a conditional vector. AttnGAN ~\cite{Tao18attngan} has also proposed a word attention model that helps model generate different images conditioned on words. As shown in Figure \ref{fig:generator}, we adapt this mechanism and an LSTM text encoder is used in our SegAttnGAN to extract word features and sentence features. The sentence feature is concatenated with random latent vector, and the word features are used as word-level attention. 

\subsection{Segmentation attention module}
The segmentation attention module is used to enhance image synthesis by persevering the spatial constraint of the input semantic maps. Park ~\etal ~\cite{park2019SPADE} has shown its efficacy and we use the same mechanism for segmentation attention module.

Mathematically, we define $F$ as features from the previous layer and $S$ as the input segmentation maps. The output of this attention module, referred from ~\cite{park2019SPADE}, to preserve spatial constrain could be expressed as in Equation \ref{attention_math}:
\begin{small}
    \begin{equation}\label{attention_math}
        \begin{aligned}
            F\prime = BN(F) * Conv(S) + Conv(S)
        \end{aligned}
    \end{equation}
\end{small}
where $BN()$ is the batch normalization function while $Conv()$ is the convolution function.

The core of this function is its property to preserve spatial information of segmentation masks. It's closely similar to the attention module in the Super Resolution task~\cite{wang2018sftgan}. By introducing the semantic map attention into each upsampling layer in a coarse-to-fine strategy, this model promisingly avoids the semantics being eliminated by pure upsampling layers.

\subsection{Segmentation mask strategies}
We have two different models when we apply different strategies for segmentation masks. The first model, named SegAttnGAN, uses pre-existing masks in the datasets as attention input. The other model, named self-attention SegAttnGAN, uses masks generated by the self-attention generator.

The self-attention generator generates segmentation masks and trained with the corresponding discriminator. Same as SegAttnGAN, it utilizes coarse-to-fine training strategy, with resolutions from 64*64, 128*128 to 256 * 256. The self-attention generator takes the same z vector and text embedding vector from SegAttnGAN as input. And at each resolution level, there is a discriminator for training.

%By fuzzing the segmentation module into the existed AttnGAN model, we utilized 2 strategies as below.

%Firstly, our SegAttnGAN takes text description as input, takes segmentation mask as attention input and generates corresponding images. As illustrated in Figure \ref{fig:attention}, we have added two attention modules with residual block after each upsample layer in generator (shown as pink connections in Figure \ref{fig:generator}). In generator, we also keep three resolution image output from 64*64, 128*128 to 256 * 256 to utilize the coarse-to-fine image GAN training. The SegAttnGAN model architecture is shown in Figure \ref{fig:generator} except "Additional Self-Attention Generator" part.

%Secondly, as shown in Figure \ref{fig:generator}, we propose the self-attention SegAttnGAN which generates segmentation mask first, then use its own generated segmentation mask as attention input to improve the image GAN training. We propose a second self-attention generator with pairing discriminator to generate segmentation mask. The second self-attention generator takes the same z vector and text embedding vector from SegAttnGAN as input, and use three resolution discriminator for low-resolution to high-resolution training. 

% \begin{figure}[h]
% \begin{center}
%     \includegraphics[width=0.9\linewidth]{latex/images/segmentation_attention.png}\end{center}
%   \caption{Detailed structure of Segmentation Attention module.}
% \label{fig:attention}
% \end{figure}
\subsection{Objective}

For the generative adversarial network, the classic objective function with conditional inputs is a min-max game between generator and discriminators defined in Equation \ref{conditional GAN}:
\begin{small}
    \begin{equation}\label{conditional GAN}
        \begin{aligned}
        \min_G \max_D V(G,D) &= E_{x\sim P_{data}(x)}[\log D(x,t)] \\
            & + E_{z\sim P_{z}(z)}[\log(1-D(G(z, t, s),t))]
        \end{aligned}
    \end{equation}
\end{small}
where $x$ refers to images from real data distribution and $z$ represents the random latent vector which drives the fake data generation. And $t$ and $s$ respectively refer to the text and segmentation input. 

Therefore, the loss function for generators is defined in Equation \ref{G loss}:
\begin{small}
    \begin{equation}\label{G loss}
        \begin{aligned}
        L_{G_i} &= -E_{z \sim P_{z}(z)}[\log(D_i(G_i(z,t,s)))] / 2 \\
            & -E_{z \sim P_{z}(z)}[\log(D_i(G_i(z,t,s),t))] / 2
        \end{aligned}
    \end{equation}
\end{small}
where the first term is an unconditional loss determining whether the image is real or fake while the second term, the conditional loss, determines whether the generated image matches the text description.

The loss function for discriminator $D_i$ is defined as in Equation \ref{D condition}:
\begin{small}
    \begin{equation}\label{D condition}
        \begin{aligned}
        L_{D_i} &= -E_{x \sim P_{data}(x)}[\log(D_i(x))] / 2 \\
        & -E_{z \sim P_{z}(z)}[\log (1-D_i(G_i(z, t, s)))] / 2 \\
        & -E_{x \sim P_{data}(x)}[\log(D_i(x,t))] / 2 \\
        & -E_{z \sim P_{z}(z)}[\log(1-D_i(G_i(z,t,s),t))] / 2
        \end{aligned}
    \end{equation}
\end{small}
where the first two terms are corresponding to the unconditional loss for optimizing discriminator while the last two terms are conditional losses.

% The overall loss function is defined in Equation \ref{total loss}:
% \begin{small}
%     \begin{equation}\label{total loss}
%         \begin{aligned}
%             L = L_{G} + \lambda L_{DAMSM}
%         \end{aligned}
%     \end{equation}
% \end{small}
% where $L_{G} = \sum_{i=0}^{m-1} L_{G_{i}}$ and $L_{DAMSM}$ follows the original DAMSM loss in \cite{Tao18attngan}

For self-attention SegAttnGAN, we define self-attention generator as $G_{s}$. We use $G_{s}(z, t)$ instead of $s$ in Equation \ref{G loss} and \ref{D condition} to define G loss and D loss. 
The overall loss is defined in Equation \ref{All loss for self attention}:
\begin{small}
    \begin{equation}\label{All loss for self attention}
        \begin{aligned}
            L = L_{G} + L_{G_s} + \lambda L_{DAMSM}
        \end{aligned}
    \end{equation}
\end{small}
where $L_{G} = \sum_{i=0}^{m-1} L_{G_{i}}$, $L_{G_s} = \sum_{i=0}^{m-1} L_{G_{s_i}}$ and $L_{DAMSM}$ follows the DAMSM loss in \cite{Tao18attngan}.
\subsection{Implementation details}
As shown in Figure \ref{fig:generator}, the generator in SegAttnGAN outputs $64*64$, $128*128$, $256*256$ images. First, we processed the segmentation mask into label maps (each channel contains different objects). And at each upsampling layer of the generator, we downsampled the segmentation label maps into the same resolution tensors as the current hidden features in the generator. Then we applied the attention module after the previous upsampling operations. The text and image encoders are following the same implementation from AttnGAN. 
%Following AttnGAN, the text encoder and image encoder stayed the same. 
For self-attention SegAttnGAN, there is no word features for the self-attention generator. The text embedding dimension is set to $256$, and loss weight $\lambda$ is set to $5.0$. ADAM solver with $beta_1=0.5$ and a learning rate of $0.0002$ are used for generator and discriminators. 

\section{Experiments}
\subsection{Dataset}
We use CUB and Oxford-102 datasets to evaluate our proposed method. CUB dataset contains images of different birds in 200 categories. We use 8841 images from this dataset for training and 2947 images for testing. Oxford-102 is another dataset consists of flower images. From this dataset, we choose 6141 images for training and 2047 images for testing.

% \begin{itemize}[parsep=10pt,itemsep=-10pt]
%     \item \textit{CUB}~\cite{WahCUB_200_2011} is an image dataset with photos of different birds in 200 categories. We are using 8841 of it for training and 2947 for testing.
%     \item \textit{Oxford-102}~\cite{Nilsback08} is a dataset of different kinds of flowers. It also provides image, segmentation and we refer from~\cite{reed2016generative} for the text description. We are using 6141 of it for training and 2047 for testing.
% \end{itemize}

\subsection{Evaluation metrics}
We use two quantitative measurements to evaluate generated images. The first metric is Inception Score ~\cite{salimans2016improved}, which has been widely used to evaluate the quality of generated images. Another metric is R-precision, which has been proposed in ~\cite{Tao18attngan} as a complimentary evaluation
metric for the text-to-image synthesis task to determine whether the generated image is well conditioned on the given text description.

\subsection{Quantitative results}
\textbf{Inception Scores:} We computed Inception Score with our generated images and compared it with those from other state-of-art methods ~\cite{reed2016generative, reed2016learning, han2017stackgan, Han17stackgan2, Tao18attngan, qiao2019mirrorgan}. The comparisons on both CUB and Oxford-102 datasets are shown in Table~\ref{tab:IS}. Our model SegAttnGAN achieves the highest Inception Score on both CUB and Oxford-102 datasets. Compared with the baseline model AttnGAN, our SegAttnGAN boosts Inception Score from $4.36$ to $4.82$ on CUB dataset. Also, our self-attention SegAttnGAN gets a good Inception Score of $4.44$ on CUB and $3.34$ on Oxford-102.
\begin{table}
\begin{center}
\begin{tabular}{|l|c|c|}
\hline
Model & CUB & Oxford-102 \\
\hline\hline
GAN-INT-CLS & 2.88$\pm$0.04 & 2.66$\pm$0.03 \\
\hline
GAWWN & 3.62$\pm$0.07 & $-$ \\
\hline
StackGAN & 3.70$\pm$0.04 & 3.20$\pm$0.01 \\
\hline
StackGAN++ & 3.82$\pm$0.06 & 3.26$\pm$0.01 \\
\hline
AttnGAN (baseline) & 4.36$\pm$0.03 & $-$ \\
\hline
MirrorGAN & 4.56$\pm$0.05 & $-$ \\
\hline
{\bf SegAttnGAN (self-attention)} & 4.44$\pm$0.06 & 3.36$\pm$0.08 \\
\hline 
{\bf SegAttnGAN} & 4.82$\pm$0.05 & 3.52$\pm$0.09 \\
\hline
\end{tabular}
\end{center}
    \caption{Inception Score of state-of-art models and our models (in bold) on CUB and Oxford-102 datasets.}
\label{tab:IS}
\vspace{-2mm}
\end{table}

\textbf{R-precision scores:} As shown in Table~\ref{tab:R}, our SegAttnGAN and self-attention SegAttnGAN also get a good R-precision score compared to AttnGAN. SegAttnGAN's score is almost the same as AttnGAN's score, indicating that SegAttnGAN can generate images consistent with input text descriptions. MirrorGAN gets the highest R-precision score as it contains a module especially for improving semantics consistency.
\begin{table}[h]
\begin{center}
\begin{tabular}{|l|c|}
\hline
Model & CUB  \\
\hline\hline
AttnGAN (baseline) & $53.31$ \\
\hline
MirrorGAN & $57.67$ \\
\hline
{\bf SegAttnGAN (self-attention)} & {\bf $52.29$} \\
\hline 
{\bf SegAttnGAN} & {\bf $52.71$} \\
\hline
\end{tabular}
\end{center}
\caption{R-precision (\%) of state-of-art models and our models.}
\label{tab:R}
\vspace{-2mm}
\end{table}

\textbf{Segmentaion attention on other models:} We also applied our segmentation attention module on StackGAN++, and the Inception scores are shown in Table~\ref{tab:IS_2}. These results indicate that our segmentation attention module can help constrain the training of different GAN models by extra semantics information and get better image generation quality.
\begin{table}[h]
\begin{center}
\begin{tabular}{|l|c|}
\hline
Model & CUB \\
\hline\hline
AttnGAN & $4.36\pm0.03$\\
\hline
{\bf AttnGAN + segmentation attention} & {\bf $4.82\pm0.05$} \\
\hline
StackGAN++ & $3.82\pm0.06$ \\
\hline 
{\bf StackGAN++ + segmentation attention} & {\bf $4.31\pm0.04$} \\
\hline
\end{tabular}
\end{center}
\caption{Inception Score comparisons of models with (in bold) and without Segmentation Attention.}
\label{tab:IS_2}
\vspace{-4mm}
\end{table}
\subsection{Qualitative results}  
\begin{figure}
\begin{center}
  \includegraphics[width=\linewidth]{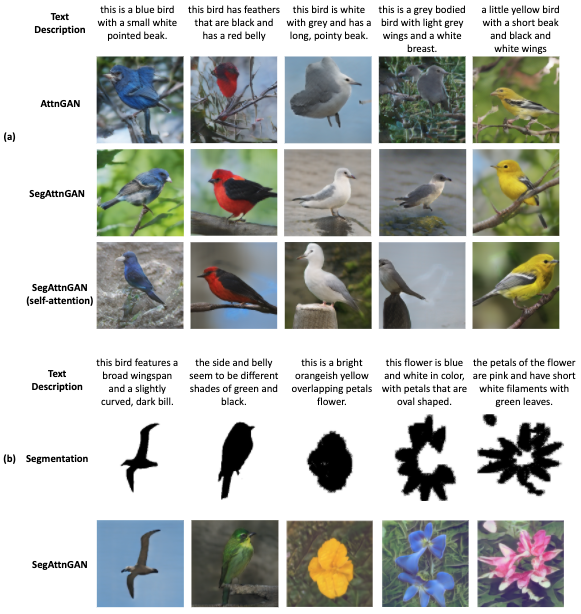}
\end{center}
   \caption{(a) Example results of our models compared to AttnGAN. (b) SegAttnGAN results with text and segmentation. }
\label{fig:result1}
\end{figure}

In Figure \ref{fig:result1}(a), we show some samples generated by AttnGAN and our models. As shown in the figure, compared to the baseline model AttnGAN~\cite{Tao18attngan}, our SegAttnGAN generates results with better object shape. Although self-attention SegAttnGAN uses generated segmentation masks, it can constrain the object shapes and generate better images than AttnGAN.
Figure \ref{fig:result1}(b) shows samples illustrating how the shape and text constrain output images of SegAttnGAN on CUB and Oxford-102 datasets. As shown in the figure, words related to color such as red and purple lead to results with different colors. The object shapes in generated images matching the input masks demonstrates that the segmentation map provides very good control over object shapes.
\subsection{Limitation and discussion}
SegAttnGAN performs well and gets the highest Inception Score compared to other methods, but this model needs segmentation input during the inference phase. Our self-attention SegAttnGAN only needs segmentation data during the training phase, and it gets better visual results compared to other models with the help of object shape constrain. But its Inception Score showed that its results get a similar image objectiveness and diversity compared to AttnGAN.
% SegAttnGAN doesn't perform promisingly on COCO dataset from both the quantitative and visualization results. Particularly, detailed information is lost in the results within the segmentation labels, that might be caused by the too strong segmentation control. Therefore, smaller weights on COCO segmentation input are required for future experiments. 
\section{Conclusion}
In this paper, we propose SegAttnGAN for text-to-image synthesis tasks, which uses segmentation attention to constrain the GAN training and successfully generates better quality images compared to other state-of-art methods. With the segmentation masks from datasets as input, our SegAttnGAN achieves the highest Inception Scores on both CUB and Oxford-102 datasets. When the masks are generated via our self-attention generator, our self-attention SegAttnGAN also generates results with better visual realism compared to other state-of-art methods.

\section{Acknowledgement}
We thank Jui-Hsin Lai and Jinghong Miao for excellent comments.

{\small
\bibliographystyle{ieee_fullname}
\bibliography{egbib}
}
\end{document}